\documentclass[letterpaper]{article} 
\usepackage{aaai24}  
\usepackage{times}  
\usepackage{helvet}  
\usepackage{courier}  
\usepackage[hyphens]{url}  
\usepackage{graphicx} 
\usepackage{natbib}  
\usepackage{caption} 
\usepackage{algorithm}
\usepackage{algorithmic}
\usepackage{booktabs}
\usepackage{multirow}
\usepackage{cleveref}

\usepackage{newfloat}
\usepackage{listings}
\DeclareCaptionStyle{ruled}{labelfont=normalfont,labelsep=colon,strut=off} 
\floatstyle{ruled}
\newfloat{listing}{tb}{lst}{}
\floatname{listing}{Listing}
\title{Dancing Avatar: Pose and Text-Guided Human Motion Videos Synthesis with Image Diffusion Model}
\author{
    Bosheng Qin\textsuperscript{\rm 1},
    Wentao Ye\textsuperscript{\rm 1},
    Qifan Yu\textsuperscript{\rm 1},
    Siliang Tang\textsuperscript{\rm 1},
    Yueting Zhuang\textsuperscript{\rm 1},
}
\affiliations{
    \textsuperscript{\rm 1}Zhejiang University\\
    $\{$bsqin, 22121058, yuqifan, siliang, yzhuang$\}$@zju.edu.cn
}

\usepackage{bibentry}

\begin{document}

\maketitle

\begin{abstract}
The rising demand for creating lifelike avatars in the digital realm has led to an increased need for generating high-quality human videos guided by textual descriptions and poses. While previous endeavors have centered on the text-to-video (T2V) diffusion model, this study delves into the untapped potential of leveraging pretrained text-to-image (T2I) models to craft human motion videos with seamless temporal coherence. We propose Dancing Avatar, designed to fabricate human motion videos driven by poses and textual cues. Our approach employs a pretrained T2I diffusion model to generate each video frame in an autoregressive fashion. The crux of innovation lies in our adept utilization of the T2I diffusion model for producing video frames successively while preserving contextual relevance. We surmount the hurdles posed by maintaining human character and clothing consistency across varying poses, along with upholding the background's continuity amidst diverse human movements. To ensure consistent human appearances across the entire video, we devise an intra-frame alignment module. This module assimilates text-guided synthesized human character knowledge into the pretrained T2I diffusion model, synergizing insights from ChatGPT. For preserving background continuity, we put forth a background alignment pipeline, amalgamating insights from segment anything and image inpainting techniques. Furthermore, we propose an inter-frame alignment module that draws inspiration from an auto-regressive pipeline to augment temporal consistency between adjacent frames, where the preceding frame guides the synthesis process of the current frame. Comparisons with state-of-the-art methods demonstrate that Dancing Avatar exhibits the capacity to generate human videos with markedly superior quality, both in terms of human and background fidelity, as well as temporal coherence compared to existing state-of-the-art approaches.

\end{abstract}

\section{Introduction}
Generating pose and text guided human motion video has imperious demand in creating avatars in the digital world. Recent advancements in the domain of pretrained text-to-image (T2I) and text-to-video (T2V) diffusion models have unfurled a panorama of ingenuity and diversity in the synthesis of images and videos. These strides have empowered users to effortlessly manifest desired characters or scenes through the simple provision of text as input \cite{RN10040, RN10128}. In recent times, a few approaches have been proposed to tackle human motion synthesis based on the given prompt and motion sequence. These approaches champion the ability to choreograph human movements in tandem with pose sequences, all the while granting control over character appearance and background through the prompts. Follow Your Pose \cite{RN10313} introduces a two-stage training paradigm to orchestrate pose-guided T2V generation via finetuning a video diffusion model. Similarly, ControlVideo \cite{RN10316} achieved text and pose-guided video synthesis with T2V diffusion model inheriting the architecture and weights from ControlNet \cite{RN10081} and the T2I diffusion model \cite{RN10040}. However, due to the limitations of modifying and fine-tuning T2I into a T2V diffusion model, the video quality of these approaches is limited and significantly lags behind the T2I diffusion model when comparing the quality of each individual frame. The manifestations of this shortcoming are evident in occurrences of incomplete characters—such as fractured hands and feet—imposing noticeable deficits in synthesis quality. Additionally, the fidelity of these approaches to maintain alignment with a given pose, and to safeguard the integrity of characters, often proves inadequate.

Diverging from earlier methodologies, our endeavor delves into the potential of leveraging a pre-trained T2I diffusion model, rather than a T2V diffusion model, to craft high-caliber and temporally consistent human motion videos. We propose Dancing Avatar, an innovative pipeline for synthesizing human motion videos correlated with text and pose sequences, capitalizing on the T2I diffusion model. This approach generates human motion videos by feeding textual descriptions of human appearance and background alongside the pose sequence. The model generates each frame of the video in a step-by-step fashion and compiles all frames to produce the final video. Navigating the challenges of employing the T2I diffusion model for human motion video synthesis entails two key hurdles. First, the challenge of sustaining uniform appearance for humans across different frames; second, preserving background consistency across frames that align with the human motion in the foreground. Addressing the initial challenge of maintaining consistent human appearance across frames, we introduce an intra-frame alignment module. This module fuses text-guided knowledge of synthesized human characters into the pretrained T2I diffusion model through fine-tuning. Our human character synthesis integrates insights from ChatGPT\footnote{https://openai.com/blog/chatgpt} and the T2I diffusion model. To ensure background consistency between synthesis frames, we introduce a background alignment pipeline. This pipeline ensures a consistent background across the human motion image sequence by leveraging knowledge from the intra-frame alignment module, segment anything \cite{RN10161}, and image inpainting techniques \cite{RN10040}. Furthermore, we propose the inter-frame alignment module to enhance the congruency of details between adjacent frames, especially the boundary between human and background. This module introduces an autoregressive character to the diffusion process, wherein the synthesis of the current image is influenced by and shares coherence with the previous frame. Comparative evaluations against state-of-the-art methods underscore the prowess of Dancing Avatar in generating human videos of superior quality, better alignment to the input prompt and poses, along with improved temporal coherence, positioning Dancing Avatar as an innovative step forward in the realm of human motion video creation.

Our main contributions can be summarized as follows:

\begin{itemize}
\item We propose Dancing Avatar, a novel pipeline for synthesizing pose and text-guided human motion videos using the T2I diffusion model.
\item We propose intra- and inter-frame alignment modules, alongside the background alignment pipeline, ensuring temporal consistency in both character and background between adjacent frames.
\item Experiment results demonstrate that Dancing Avatar can synthesize human motion videos with significantly superior quality, alignment with input conditions, and temporal consistency compared to previous state-of-the-art approaches.
\end{itemize}

\section{Related Work}
\subsection{Controllable Image and Video Synthesis}
Controllable image synthesis has evolved from early text-controlled approaches like DALLE·2 \cite{RN10085} and Stable Diffusion \cite{RN10040} to more sophisticated methods that enable fine-grained control. Methods such as Prompt-to-Prompt \cite{RN10041}, InstructPix2Pix \cite{RN10108}, ChatGenImage \cite{RN10331}, and HIVE \cite{RN10151} employ image editing techniques, while others like ControlNet \cite{RN10081} and T2I-Adapter \cite{RN10330} utilize conditional networks to extend pre-trained T2I diffusion models. Some works achieve controllable image synthesis through model finetuning, connecting image concepts with word embeddings, as in DreamBooth \cite{RN10011}, Textual Inversion \cite{RN10015}, and Custom Diffusion \cite{RN10042}.

In the realm of controllable video synthesis, advances mirror those in image generation. Approaches like Tune-A-Video \cite{RN10117}, Video-P2P \cite{RN10116}, and ControlVideo \cite{RN10343} edit given videos through T2V diffusion model fine-tuning. Meanwhile, zero-shot editing, exemplified by Vid2Vid Zero \cite{RN10176}, VidEdit \cite{RN10340}, and InstructVid2Vid \cite{RN10315}, leverages pre-trained T2I visual knowledge for editing unseen samples.

\subsection{Human Motion Video Synthesis}
Early human video synthesis utilized LSTM \cite{RN10297} and GANs \cite{RN10305, RN10288, RN10287, RN10314}. Vid2Vid \cite{RN10287} approached motion transfer with conditional GANs \cite{RN841}, while Few-shot Vid2Vid \cite{RN10305} achieved few-shot generalization through a weight generation module. DisCo \cite{RN10317} introduced disentangled control in reference-based human motion video synthesis, predicting frames based on specific attributes.

However, these methods require character-specific fine-tuning, limiting motion synthesis to targeted characters. Recent approaches like Follow Your Pose \cite{RN10313} and ControlVideo \cite{RN10316} use T2V diffusion models for text and pose-guided synthesis. In contrast, emerging methods yield better flexibility, allowing customization of character and background appearance through text instructions.

Similar to Follow Your Pose \cite{RN10313} and ControlVideo \cite{RN10316}, Dancing Avatar undertakes pose and text-guided human motion video synthesis. Distinguishing itself, Dancing Avatar harnesses pre-trained T2I diffusion models instead of T2V diffusion models to synthesize temporally coherent video. This strategy enhances video quality, alignment to input conditions, and inter-frame correlation, resulting in coherent, appealing outcomes.

\section{Method}
Our approach begins with an overview of the Dancing Avatar pipeline. We then delve into the specifics of the intra-frame alignment module, responsible for harmonizing human appearances across frames with varying pose conditions. Proceeding, we outline the background alignment pipeline. Furthermore, we introduce the inter-frame alignment module, which employs an auto-regressive pattern to enhance the alignment of details between adjacent frames.

\begin{figure*}
\centering
\includegraphics[width=\textwidth]{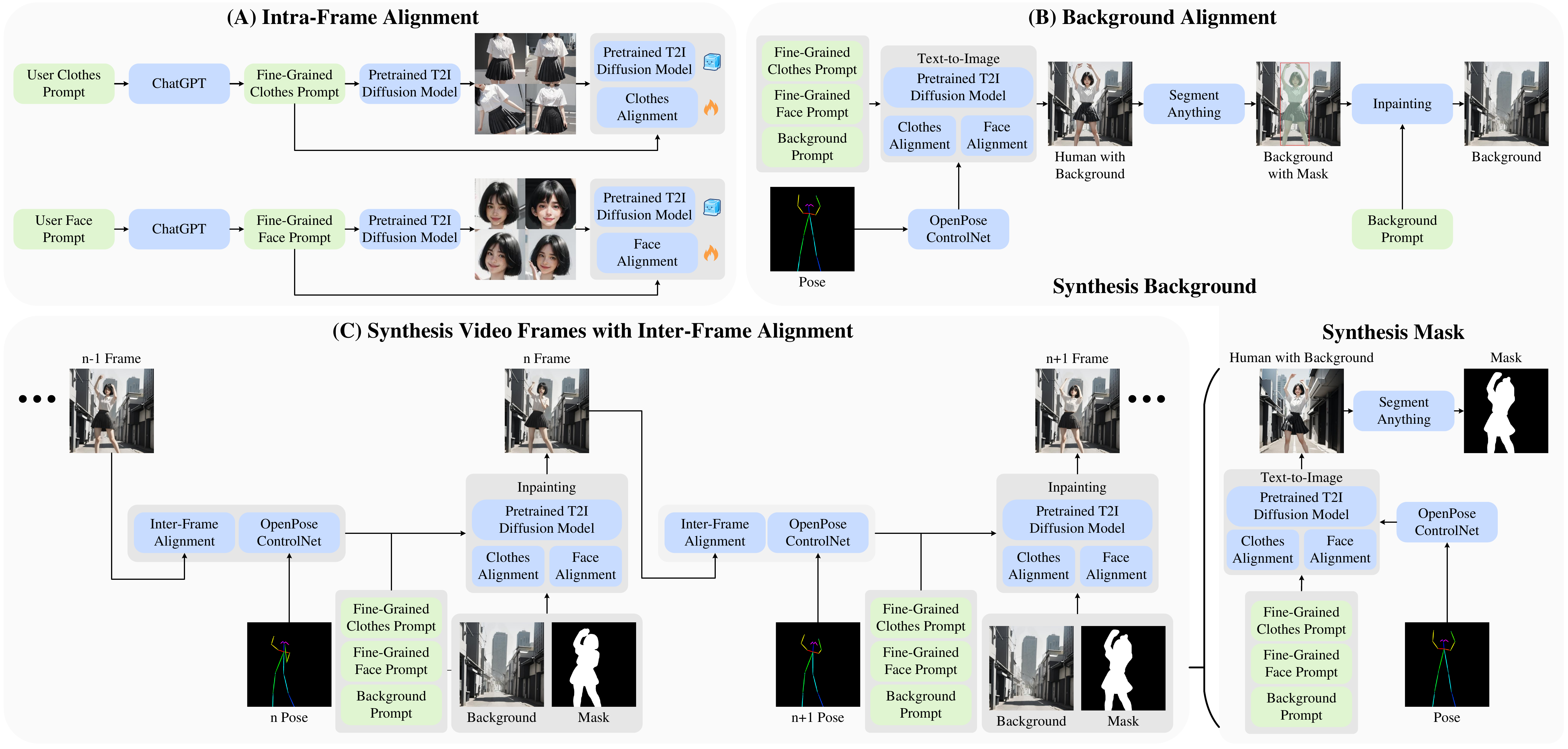}
\caption{Pipeline of Dancing Avatar. The Dancing Avatar pipeline begins by acquiring the intra-frame alignment module. It then executes the background alignment process. Ultimately, it synthesizes each frame of the human motion video in an auto-regressive fashion, assisted by the inter-frame alignment module.}
\label{Dancing Avatar}
\end{figure*}

\subsection{Overview of Dancing Avatar}
Dancing Avatar leverages the T2I diffusion model to create pose and text-guided human motion videos. Given textual prompts encompassing [clothes, face, background], the pipeline generates the videos by concatenating images produced under consistent textual conditions but varying pose conditions using an autoregressive approach. The pipeline maintains human and background coherence across frames through the intra-frame alignment module, the inter-frame alignment method, and the background alignment pipeline, as illustrated in Figure \ref{Dancing Avatar}.

The pipeline of Dancing Avatar starts with obtaining the intra-frame alignment module. The intra-frame alignment module is fine-tuned by integrating ChatGPT's knowledge and a pre-trained T2I diffusion model. ChatGPT aids in refining user-provided coarse-grained clothing and facial prompts into more detailed ones. The intra-frame alignment module collaborates with the pre-trained T2I diffusion model to ensure uniformity in human clothing and facial appearance across different frames.

Subsequently, Dancing Avatar generates contextually appropriate background images for the human motion videos, adhering to textual specifications through segment anything and image inpainting techniques. Moreover, distinct pose-guided images are synthesized for different poses, yielding human area masks. This process culminates in obtaining background images and human masks for each pose.

Then the Dancing Avatar start to synthesis human motion video with fine-grained prompt and different pose, where ControlNet \cite{RN10081} manages pose inputs. Accompanied by the previously obtained background image and human mask, the pipeline employs inpainting to restore the masked area according to the given pose and prompt. Throughout this process, the intra-frame alignment module preserves consistency in human clothing and facial appearance. The inter-frame alignment module, utilizing the previous synthesis video as input, further enhances detailed coherence between adjacent frames, especially the boundary between human and background. Concatenating these pose-varied images ultimately produces the pose and text-guided human motion video.

\subsection{Intra-Frame Alignment}
Dancing Avatar employs a concatenation of image sequences from the T2I diffusion model to create human motion videos. A central challenge involves preserving human clothing and facial appearances across images synthesized with varying poses, guided by the [clothes, face] prompt. In addressing this, we introduce the intra-frame alignment module, which harnesses ChatGPT's visual knowlege to augment the pre-trained T2I diffusion model.

ChatGPT, although primarily linguistic, excels in generating finely detailed image descriptions. Leveraging this, we explore ChatGPT's potential in refining coarse-grained [clothes, face] prompts into fine-grained descriptions. These fine-grained descriptions precisely control the pre-trained T2I diffusion model to synthesize images featuring consistent clothing or facial attributes.

For precise clothing descriptions, we prompt ChatGPT with "Imagine an image with [clothes], and describe the attribute of the clothes in 20 single words with detailed precision, making sure that the words correspond to one and only one clothes only." This process yields detailed clothing descriptions from the given coarse [clothes] description. Similarly, we prompt ChatGPT with "Imagine the [face] and describe it with 20 single words, focusing on facial and hair details, making sure that the description of the face and hair corresponds to the face and hair of one and only one person." This results in fine-grained facial appearance descriptions. The individual fine-grained clothing and facial prompts are then provided separately to the pre-trained T2I diffusion model, producing two sets of images: one with consistent clothing and another with consistent facial appearances.

These image sets undergo training to develop the clothes and face intra-frame alignment modules, responsible for clothing and facial appearance control, respectively. Following \cite{RN10103}, the alignment module adopts the architecture of the original cross-attention layer from the pre-trained T2I diffusion model, enhanced with low-rank optimization for efficiency. Internal features from each cross-attention layer in the alignment module are incorporated into the corresponding layer of the UNet \cite{RN10137} within the T2I diffusion model to introduce specialized knowledge. During training, the original T2I diffusion model remains fixed, we exclusively optimize the intra-frame alignment module. The T2I diffusion model with the intra-frame alignment module receives the previously mentioned synthesized clothing or facial images, paired with fine-grained prompts from ChatGPT. Post-training, distinct clothes and facial intra-frame alignment modules emerge, seamlessly integratable into the pre-trained T2I diffusion model to synthesize images featuring consistent human appearances when provided with the same fine-grained prompt.

\subsection{Background Alignment}
The second challenge revolves around maintaining consistent backgrounds within an image sequence that features varying human poses. Here, we introduce the background alignment pipeline, encompassing background and human mask synthesis. This process, when integrated with human synthesis corresponding to various pose conditions, facilitates seamless pose-specific human insertion into a consistent background within the synthesized video.

In the initial stage, we synthesize a background image tailored to the human pose video while adhering to specified text requirements. Through the T2I diffusion model and the intra-frame alignment module, we synthesize a human character image under the prescribed background by utilizing textual descriptions. Subsequently, employing the segment anything \cite{RN10161}, we derive a body area mask. Leveraging this mask, we perform inpainting to seamlessly integrate the background condition with the human-free area, yielding a background image devoid of any human presence. This image serves as a fitting backdrop for the human video.

The subsequent step involves generating human masks specific to the target pose sequence. To achieve this, we synthesize the desired human character using the T2I diffusion model and the intra-frame alignment module, employing the target pose while keeping the background prompt constant. Utilizing segment anything, we extract the human mask aligned with the target pose condition.

With both the background image and the sequence of target pose masks in hand, we proceed to inpainting. Guided by the background image and human area mask, we inpaint the masked regions based on the pose sequence and input prompts. This final step results in a collection of images, each featuring distinct human poses set against a uniform background. Concatenating these images generates a seamless human motion video. This process ensures a consistent background while accommodating variations in human poses.

\begin{table*}[t]
\caption{Comparison of video quality between Dancing Avatar and previous state-of-the-art approaches. The best model is made bold.}
\label{table1}
\begin{center}
\resizebox{\textwidth}{!}{
\begin{tabular}{ccccccc}
\toprule
Model&  Frame NIQE  $\downarrow$&	Body NIQE  $\downarrow$&	Background NIQE  $\downarrow$&	Frame BRISQUE  $\downarrow$&	Body BRISQUE  $\downarrow$&	Background BRISQUE  $\downarrow$ \\
\midrule 
Follow Your Pose \cite{RN10313}&
5.27&	9.13&	4.18&	36.67&	47.89&	46.93	\\
ControlVideo \cite{RN10316} &
3.32&	9.01&	3.21&	26.21&	48.11&	48.11	\\
\textbf{Dancing Avatar}&
\textbf{2.99}&	\textbf{5.03}&	\textbf{2.44}&	\textbf{19.56}&	\textbf{45.19}&	\textbf{43.75}	\\
\bottomrule
\end{tabular}}
\end{center}
\end{table*}

\begin{table*}[t]
\caption{Comparison of input alignment and temporal consistency between adjacent frames between Dancing Avatar and previous state-of-the-art approaches. The best model is made bold.}
\label{table2}
\begin{center}
\resizebox{\textwidth}{!}{
\begin{tabular}{ccc|ccccc}
\toprule
\multirow{2}{*}{Model}& \multicolumn{2}{c}{Input Alignment}&   \multicolumn{5}{c}{Frame Consistency}\\
&   Pose MES  $\downarrow$&  Text Alignment $\uparrow$& Frame MSE  $\downarrow$&	Frame L1  $\downarrow$&	Frame CLIP $\uparrow$&	Body CLIP $\uparrow$&	Background CLIP $\uparrow$\\
\midrule 
Follow Your Pose \cite{RN10313}&
10.76&	27.63&   97.54&	364.72&	74.64&	71.24&	71.22 \\
ControlVideo \cite{RN10316} &
30.57&	27.14& 50.15&	362.19&	72.60&	73.29&	76.46\\
\textbf{Dancing Avatar}&
\textbf{1.48}&	\textbf{31.92}&    \textbf{26.66}&	\textbf{270.31}&	\textbf{80.63}&	\textbf{77.43}&	\textbf{81.82} \\
\bottomrule
\end{tabular}}
\end{center}
\end{table*}

\subsection{Inter-Frame Alignment}
While we've introduced the intra-frame alignment module and the background alignment pipeline to ensure consistency in the synthesized video's human characters and background, certain inconsistencies, particularly along the boundaries between human and background in adjacent frames during inpainting, may persist. In this section, we introduce the inter-frame alignment, which empowers the T2I diffusion model to auto-regressively synthesize a sequence of human motion frames, enhancing the coherence of details between consecutive frames.

The inter-frame alignment module strives to enhance consistency between adjacent frames by utilizing the prior frame as a guiding input for the current frame's synthesis. This module adopts the encoder block architecture of U-Net \cite{RN10137}. Following \cite{RN10081}, the module's features are directly incorporated into the intermediate features of the original U-Net, serving as a conditional input while retaining the pre-trained parameters. Training the inter-frame alignment module mandates an extensive collection of human motion videos. These videos are fragmented into frames, each associated with a caption from the BLIP-2 model \cite{RN10345}. During training, the pre-trained T2I diffusion model remains static, and the sole optimization occurs within the inter-frame alignment module. The optimization objective entails accurately predicting the present frame based on the preceding frame and textual description. Upon training completion, the inter-frame alignment module discerns correlations between neighboring frames. During inference, Dancing Avatar sequentially synthesizes frames in an auto-regressive manner, generating images according to pose chronology. The preceding synthesized image acts as input for the inter-frame alignment module, bolstering consistency between adjacent frames.

\begin{table*}[t]
\caption{Ablation experiment about the video synthesis performance of Dancing Avatar in terms of consistency between adjacent frames and video quality.}
\label{table3}
\begin{center}
\resizebox{\textwidth}{!}{
\begin{tabular}{cccccc|cc}
\toprule
\multirow{2}{*}{Model}& \multicolumn{5}{c}{Frame Consistency}& \multicolumn{2}{c}{Quality} \\
& Frame MSE $\downarrow$&	Frame L1 $\downarrow$&	Frame CLIP $\uparrow$&	Body CLIP $\uparrow$&	Background CLIP $\uparrow$&	BRISQUE $\downarrow$&	NIQE $\downarrow$ \\
\midrule 
w/o intra-frame alignment &
27.31&	308.79&	78.76&	75.66&	82.00&	19.65&	3.05 \\
w/o inter-frame alignment &
26.94&	281.85&	79.41&	76.40&	81.14&	19.58&	3.01 \\
w/o background alignment &
117.79&	376.13&	78.56&	77.73&	79.92&	19.51&	2.93 \\
\textbf{Dancing Avatar}&
26.66&	270.31&	80.63&	77.43&	81.82&	19.56&	2.99 \\
\bottomrule
\end{tabular}}
\end{center}
\end{table*}

\begin{figure*}
\centering
\includegraphics[width=\textwidth]{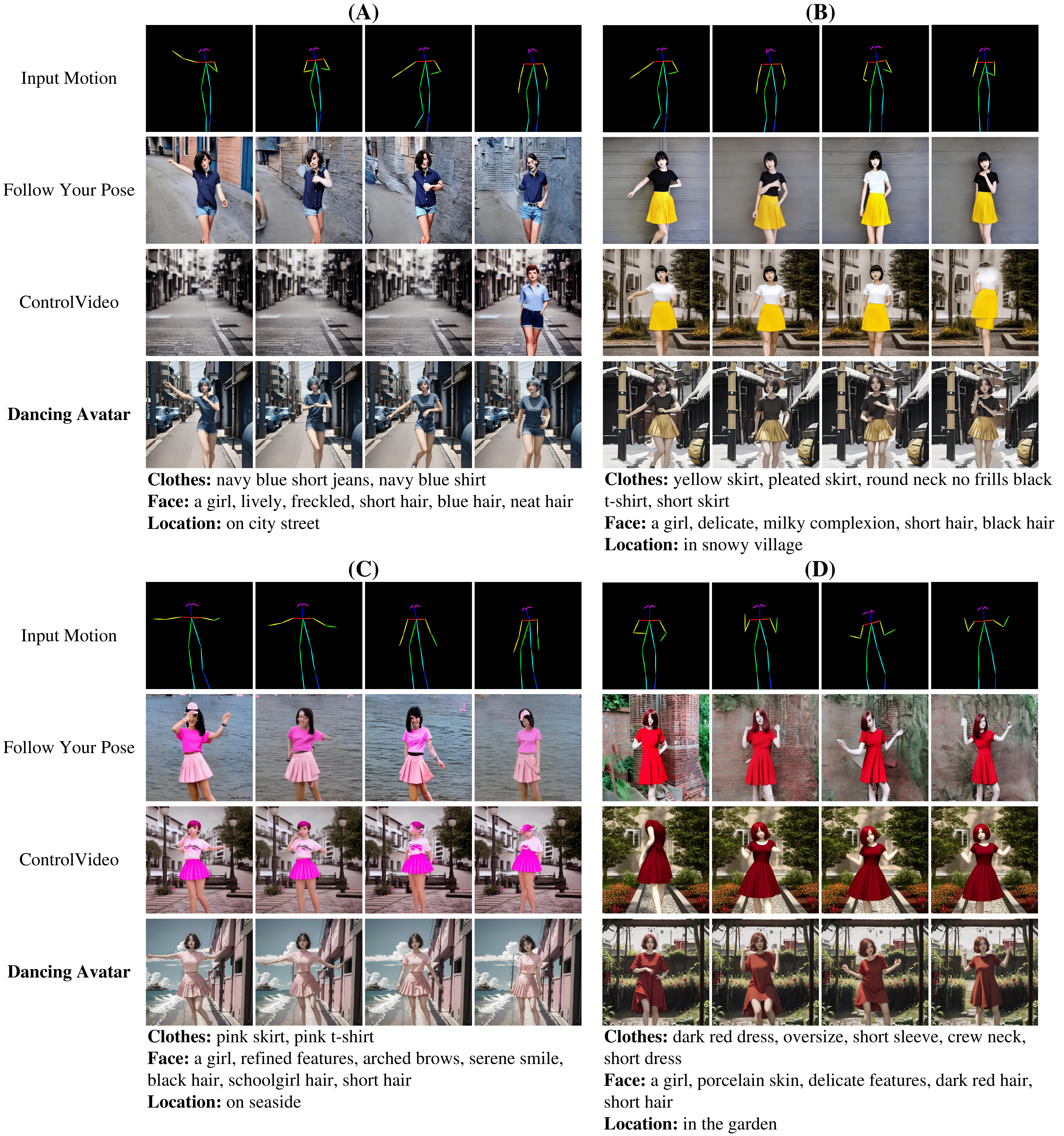}
\caption{Comparison between Dancing Avatar and previous state-of-the-art pose and text-guided human motion video synthesis approaches. The videos synthesized by Dancing Avatar exhibit higher video quality, alignment with inputs, and temporal coherence maintained across successive frames.}
\label{Dancing Avatar Comaprsion}
\end{figure*}

\section{Experiments}
In this section, we present a comprehensive range of qualitative and quantitative experiments that showcase the remarkable prowess of Dancing Avatar. We will clearly demonstrate how Dancing Avatar surpasses the benchmarks set by prior state-of-the-art methods, excelling in aspects such as video quality, alignment with inputs, and the temporal coherence maintained across successive frames. Furthermore, we conduct ablation experiments to highlight the individual contributions of the intra-frame alignment module, inter-frame alignment module, and background alignment pipeline. For a comprehensive understanding of our experimental setup, implementation particulars, and the dataset employed, we refer readers to the Appendix.

\subsection{Evaluation Matrices}
Thirteen diverse evaluation metrics were employed to comprehensively assess video quality, alignment with inputs, and temporal coherence between consecutive frames. BRISQUE \cite{RN10324} and NIQE \cite{RN10323} gauged frame quality, encompassing overall, human-part, and background-part aspects. Pose accuracy was evaluated using MSE, while CLIP text-image similarity measured alignment with the prompt. Temporal consistency between frames was quantified using L1, MSE, and CLIP \cite{RN5621}. Furthermore, the distance between human and background segments in CLIP evaluated the integrity of respect parts. Additional insights into these evaluation matrices' implementation are provided in the Appendix.

\subsection{Qualitative Comparision with State-of-the-Arts}
We visually contrast the synthesized videos from Dancing Avatar with state-of-the-art text and pose-controlled video synthesis baselines, including Follow Your Pose \cite{RN10313} and ControlVideo \cite{RN10316}, as shown in Figure \ref{Dancing Avatar Comaprsion}.

Dancing Avatar notably elevates frame quality by capturing intricate person and background details. In Figure \ref{Dancing Avatar Comaprsion} (A), for instance, Dancing Avatar adeptly maintains arm continuity, facial proportions, and background fidelity, in contrast to Follow Your Pose. Furthermore, Dancing Avatar generates more lifelike scenes with improved lighting and object integration compared to ControlVideo, illustrated in Figure \ref{Dancing Avatar Comaprsion} (D).

Dancing Avatar establishes a commendable alignment with both pose and text inputs. In Figure \ref{Dancing Avatar Comaprsion} (A), only Dancing Avatar faithfully synthesizes the intended initial pose. Dancing Avatar also excels in adhering to text conditions; as shown in Figure \ref{Dancing Avatar Comaprsion} (B), only Dancing Avatar correctly renders the snowy village background.

Temporal consistency across consecutive frames emerges as another distinct strength of Dancing Avatar. While Follow Your Pose (Figure \ref{Dancing Avatar Comaprsion} B) showcases abrupt clothing changes, Dancing Avatar upholds consistent clothing attributes. Additionally, it ensures consistent facial appearances, avoiding abrupt hairstyle alterations evident in ControlVideo (Figure \ref{Dancing Avatar Comaprsion} C). Background coherence is also heightened in Dancing Avatar; it sustains the backdrop scene, avoiding drastic alterations evident in Follow Your Pose (Figure \ref{Dancing Avatar Comaprsion} D) and global color shifts as observed in Follow Your Pose (Figure \ref{Dancing Avatar Comaprsion} C).

\subsection{Quantitative Comparision with State-of-the-Arts}
We perform quantitative comparisons between Dancing Avatar and previous state-of-the-art methods in terms of video quality, input alignment, and temporal consistency, where Dancing Avatar excels in all three dimensions.

Table \ref{table1} presents the video quality evaluation. We utilize NIQE and BRISQUE metrics to assess models, considering overall frame quality, body segment quality, and background segment quality. Across all these aspects, Dancing Avatar significantly surpasses Follow Your Pose and ControlVideo. Notably, it achieves the lowest Frame, Body, and Background NIQE scores, indicating its proficiency in generating high-quality frames for humans and backgrounds. Additionally, the model achieves the lowest Frame, Body, and Background BRISQUE scores, affirming its capacity to produce visually captivating and authentic video content for both human and background segments.

Table \ref{table2} presents a detailed comparison of input alignment and temporal consistency between adjacent frames. For input alignment, we measure MSE between input and synthesis poses, while CLIP evaluates text-to-video frame alignment. Dancing Avatar excels in maintaining precise input alignment, yielding the lowest Pose MSE and highest Text Alignment values among all methods. 

Temporal consistency highlights the Dancing Avatar superiority even further. With the lowest Frame MSE and Frame L1 values, Dancing Avatar demonstrates its knack for ensuring consistent motion and content across frames. Moreover, it secures the highest Frame, Body, and Background CLIP image-to-image scores, accentuating its proficiency in preserving both global and local content cohesion over time.

\begin{figure}[t]
\centering
\includegraphics[width=0.47 \textwidth]{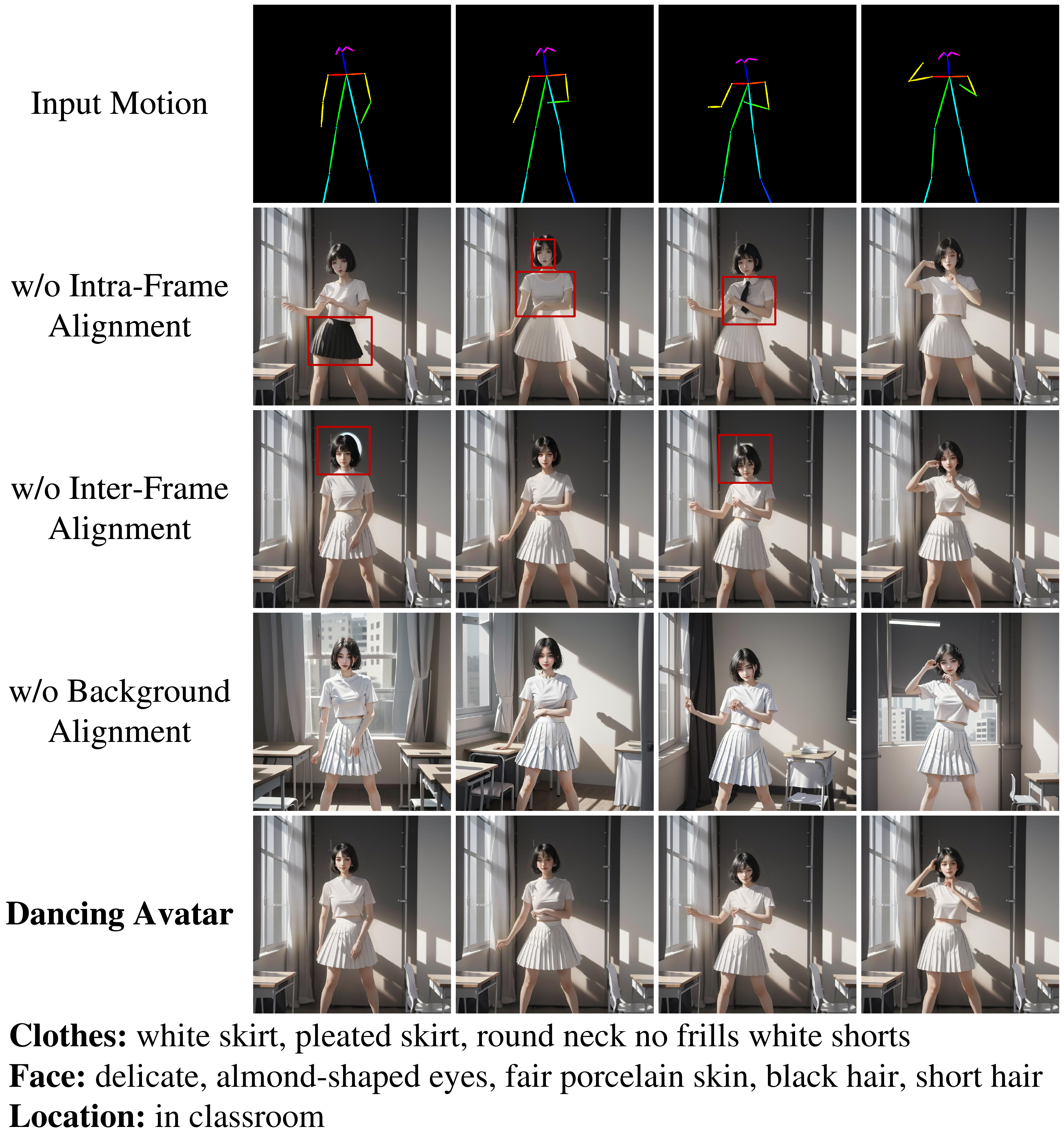}
\caption{Visualized result of ablation experiment. Dancing Avatar demonstrates notably greater consistency between adjacent frames in comparison to its ablation counterparts.}
\label{Ablation}
\end{figure}

\subsection{Ablation Study}
In this section, we conduct ablation experiments to comprehensively evaluate the distinct contributions of three pivotal modules within the Dancing Avatar framework: intra-frame alignment, inter-frame alignment, and background alignment. We will demonstrate that all three modules substantially contribute to maintaining video consistency across adjacent frames while upholding video quality.

\textbf{Intra-Frame Alignment.} The intra-frame alignment module aims to ensure consistent human appearance across diverse frames. We compare the synthesized human motion video from Dancing Avatar with the intra-frame alignment module against ablation counterparts utilizing only fine-grained human character descriptions from ChatGPT as direct text input. These fine-grained descriptions represent an optimal prompt that could be provided to Dancing Avatar without the intra-frame alignment module. The visual results in Figure \ref{Ablation} underscore that Dancing Avatar with the intra-frame alignment module exhibits greater consistency in human appearance between adjacent frames compared to the approach relying solely on ChatGPT descriptions. Notably, in the ablation group without the intra-frame alignment module, both clothing and human appearances exhibit discontinuities across frames. This discrepancy mainly arises due to the varying strengths of different T2I diffusion models. We use different T2I diffusion models to synthesize clothes, facial, and human pose image sequences. Those adept at synthesizing final human videos with backgrounds but less proficient in capturing clothing or facial details.

Quantitative experiments reveal a significant reduction in Body CLIP scores without the intra-frame alignment module. This indicates that the absence of intra-frame alignment leads to less coherent content across frames, resulting in diminished alignment of human appearance. Additionally, the slight decrease in video quality of Dancing Avatar, as measured by BRISQUE and NIQE, when incorporating the intra-frame alignment module suggests its contribution to slight improvements in video quality.

\textbf{Inter-Frame Alignment} The inter-frame alignment module targets enhanced detail alignment among adjacent frames, particularly at the boundary between human and background. As evident in Figure \ref{Ablation}, Dancing Avatar without the inter-frame alignment module displays abrupt lighting changes in the boundary area of the first and third frames. Quantitative analysis in Table \ref{table3} indicates a marginal increase in Frame MSE and Frame L1 scores when omitting the inter-frame alignment module. This underlines the module's role in preserving overall scene integrity.

\textbf{Background Alignment} The Background Alignment pipeline ensures consistent backgrounds across distinct human poses. Visual observations in Figure \ref{Ablation} demonstrate that the absence of the background alignment pipeline leads to abrupt background changes between adjacent frames, even when the same background prompt is provided. Quantitative analysis echoes these findings, revealing a substantial decline in Background CLIP scores in Table \ref{table3}. This emphasizes the background alignment module's critical role in maintaining background consistency.

\section{Conclusion}

We present Dancing Avatar, a groundbreaking framework for generating high-quality human motion videos guided by poses and textual cues. By leveraging the capabilities of pretrained text-to-image (T2I) models, the proposed method transcends the limitations of conventional text-to-video (T2V) diffusion models, producing human videos of remarkable quality and temporal coherence. Comparative evaluations against existing state-of-the-art methods showcase the superior quality of human and background fidelity, as well as enhanced temporal consistency offered by Dancing Avatar.

\bibliography{aaai24, my_endnote_cite, conference_cite}

\clearpage
\appendix
\section{Appendix}

\subsection{Implemetation Details}
Our experimentation utilizes PyTorch \cite{RN9728}, Hugging Face Diffusers\footnote{https://github.com/huggingface/diffusers}, and Numpy \cite{RN3304}. For the intra-frame alignment module, we exploit Diffusion Fashion\footnote{https://huggingface.co/MohamedRashad/diffusion$\_$fashion} for clothing synthesis and MajicMIX-Realistic-v6 Diffusion\footnote{https://huggingface.co/digiplay/majicMIX$\_$realistic$\_$v6} for facial appearance. Both T2I diffusion models follow the structure of Stable Diffusion \cite{RN10040}. By employing yolo-v3 \cite{RN10346}, we detect and crop facial and clothing areas. The cropped images are utilized for training the respective intra-frame alignment modules. Training entails 500 steps per module, employing the AdamW optimizer \cite{RN10327}. The inter-frame alignment module is trained over 20 epochs with a learning rate of 5e-5, also using the AdamW optimizer. During video synthesis, we implement the DPM adaptive sampler \cite{RN10326} with 25 diffusion steps. ControlNet-v1.1\footnote{https://huggingface.co/lllyasviel/ControlNet-v1-1} with OpenPose \cite{RN10081, RN10328} allows the MajicMIX-Realistic-v5 Diffusion\footnote{https://huggingface.co/digiplay/majicMIX$\_$realistic$\_$v5} to synthesize human poses as required. We set the text guidance scale to 7.5 for the synthesis process.

\subsection{Datasets}
Dancing Avatar draws on videos to train the inter-frame alignment module and relies on pose sequences during inference.

Human Motion Data: We gather 1162 videos showcasing human motion, each exceeding a resolution of 512*512 and resampled to 24 fps. Adjacent frames are paired [current frame, next frame, caption], where the caption originates from the BLIP-2 model \cite{RN10345}, representing the image caption of [next frame]. During training, the inter-frame alignment module predicts the forthcoming frame based on the current frame and its associated caption.

Sequence of Pose Data: This input is pivotal for Dancing Avatar during inference. We compile 10 brief dancing videos of human subjects from the web, each cropped into 1.5-second segments. OpenPose \cite{RN10328} detects the human poses within these segments, resulting in the required pose maps.

\subsection{Evaluation Matrices}
We use thirteen distinct evaluation metrics to assess video quality, alignment with inputs, and temporal consistency among frames. BRISQUE \cite{RN10324} and NIQE \cite{RN10323} analyze frame quality. The matrices provide insights into global frame quality, human-part quality, and background-part quality. Pose accuracy is evaluated using MSE, while CLIP measures alignment between synthesized video and prompt. For temporal consistency, we use L1, MSE, and CLIP \cite{RN5621}. Human and background part consistency between frames is measured using CLIP distance.

\textbf{Blind/Referenceless Image Spatial Quality Evaluator (BRISQUE) \cite{RN10324}.} We use BRISQUE to evaluate the quality of the synthesized video. BRISQUE analyzes the spatial properties of an image to estimate its quality. The image is first divided into non-overlapping square patches, and statistical properties are calculated for each patch. These properties are then forwarded to a pre-trained model to predict image quality. In our experiment, we extract every frame from the synthesized video and calculate the mean BRISQUE score for all frames as the Frame BRISQUE score. Additionally, we extract the human area by segmenting individuals from every frame in the synthesized video and calculate the Body BRISQUE score. We also extract the background area and calculate the Background BRISQUE score. Lower BRISQUE values indicate higher image quality.

\textbf{Naturalness Image Quality Evaluator (NIQE) \cite{RN10323}.} We use NIQE to evaluate the quality of the synthesized video. NIQE measures the deviation of an image from the statistics of natural images, thus estimating its perceived quality. Similar to the BRISQUE experiment, we report the mean NIQE score for every frame in the synthesized video as the Frame NIQE score. We segment the body and background areas and report the Body NIQE and Background NIQE scores, respectively. A lower NIQE score suggests that the frame closely resembles the statistical properties of a natural image, indicating higher video quality.

\textbf{Pose Mean Squared Error (MSE).} Pose MSE is utilized to assess the fidelity of the synthesized human poses, offering insights into the model's ability to accurately reproduce the intended movements from the provided input. We calculate the MSE between the given pose and the pose from the synthesized frames extracted with OpenPose \cite{RN10328}. A lower score indicates that the poses in the synthesized frames are more aligned with the given poses.

\textbf{CLIP \cite{RN5621} Text Consistency.} We use CLIP Text Consistency to evaluate the textual alignment of the generated video with the given prompts. We extract every frame from the synthesized video, and the mean value of text-image similarities in the CLIP feature space is calculated as text consistency. A higher score indicates that the synthesized video is more semantically related to the given prompt.

\textbf{L1 Frame Consistency.} We calculate the L1 difference in pixel level of all adjacent frames in the videos to measure temporal consistency. A lower value indicates more similarity in adjacent frames.

\textbf{MSE Frame Consistency.} We calculate the MSE loss in pixel level of all adjacent frames in the videos to measure temporal consistency. A lower value indicates more similarity in adjacent frames.

\textbf{CLIP \cite{RN5621} Frame Consistency.} We calculate CLIP image-to-image similarity to evaluate the temporal consistency of the generated video. We use the CLIP image encoder to extract features from the generated video and compute the cosine similarity between adjacent frames, reporting the Frame CLIP score. Additionally, we report the CLIP similarity between adjacent frames of body and background segments and present the Body and Background CLIP scores. A higher score indicates that the synthesized video shares more temporal consistency between adjacent frames.

\end{document}